\documentclass[a4paper,conference]{IEEEtran}
\IEEEoverridecommandlockouts
\usepackage{ifpdf}

\ifCLASSINFOpdf
   \usepackage[pdftex]{graphicx}
\else
   \usepackage[dvips]{graphicx}
\fi

\usepackage{cite}
\usepackage{amsmath}
\usepackage{algorithmic}
\usepackage{array}
\usepackage{url}
\usepackage{pifont}
\usepackage{multirow}
\ifCLASSOPTIONcompsoc
 \usepackage[caption=false,font=normalsize,labelfont=sf,textfont=sf]{subfig}
\else
 \usepackage[caption=false,font=footnotesize]{subfig}
\fi

\begin{document}

\title{Self-Training for Domain Adaptive Scene Text Detection}

\author{\IEEEauthorblockN{Yudi Chen\textsuperscript{1}\textsuperscript{2}\IEEEauthorrefmark{1},
Wei Wang\textsuperscript{1}\textsuperscript{2}\IEEEauthorrefmark{1},
Yu Zhou\textsuperscript{1}\IEEEauthorrefmark{2},
Fei Yang\textsuperscript{3},
Dongbao Yang\textsuperscript{1},
Weiping Wang\textsuperscript{1}}
\IEEEauthorblockA{\textsuperscript{1}Institute of Information Engineering, Chinese Academy of Sciences}
\IEEEauthorblockA{\textsuperscript{2}University of Chinese Academy of Sciences}
\IEEEauthorblockA{\textsuperscript{3}TAL Education Group}
\thanks{\IEEEauthorrefmark{1}Equal contribution}
\thanks{\IEEEauthorrefmark{2}Corresponding author, email: zhouyu@iie.ac.cn}}

\maketitle
\begin{abstract}
   Though deep learning based scene text detection has achieved great progress, well-trained detectors suffer from severe performance degradation for different domains. In general, a tremendous amount of data is indispensable to train the detector in the target domain. However, data collection and annotation are expensive and time-consuming. To address this problem, we propose a self-training framework to automatically mine hard examples with pseudo-labels from unannotated videos or images. To reduce the noise of hard examples, a novel text mining module is implemented based on the fusion of detection and tracking results. Then, an image-to-video generation method is designed for the tasks that videos are unavailable and only images can be used. Experimental results on standard benchmarks, including ICDAR2015, MSRA-TD500, ICDAR2017 MLT, demonstrate the effectiveness of our self-training method. The simple Mask R-CNN adapted with self-training and fine-tuned on real data can achieve comparable or even superior results with the state-of-the-art methods. 
\end{abstract}

\section{Introduction}


With the development of deep learning, remarkable progress has been made in the area of scene text detection in supervised learning scenario \cite{he2017deep,liao2017textboxes,liu2017deep,liu2019towards,long2018textsnake,tian2016detecting,Wang2019shape,ye2015text,zhou2017east}. However, this requires that the training data and the test data come from the same domain, which is not always the case in real-world scenarios. For a specific application task, a new training dataset must be collected and annotated, which is very expensive and time-consuming. In recent years, several studies including weakly/semi-supervised learning, data generation, and domain adaptation have been proposed to solve this problem.



\begin{figure}[t]
\centering
\includegraphics[width=1\columnwidth]{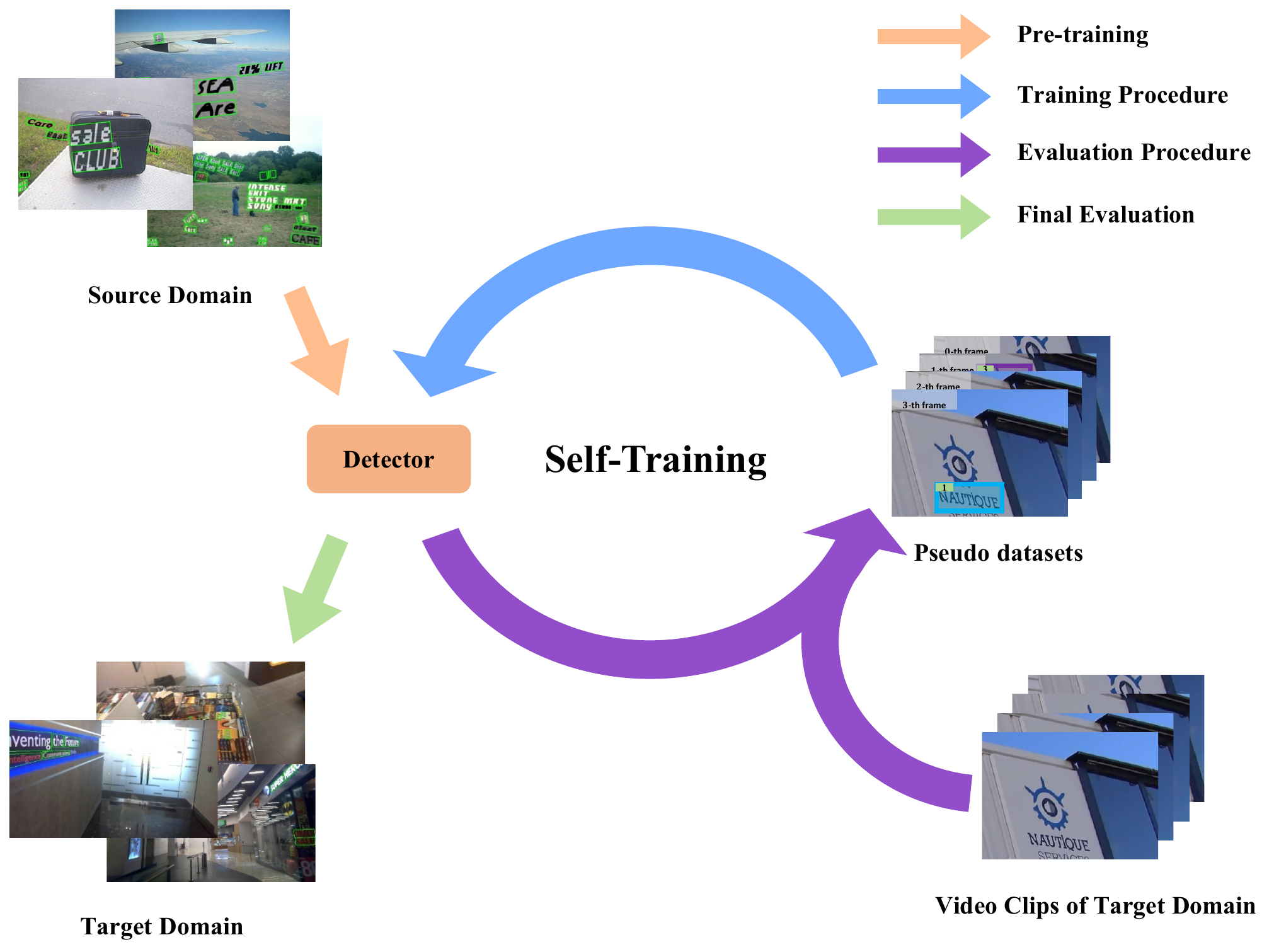}
\caption{Cross-domain text detection by the self-training framework. First, the detector is initialized by the annotated data from the source domain. Then videos and images from the target domain are evaluated by the detector to generate the pseudo datasets. After data cleaning and hard example mining, the pseudo datasets are used to re-train the detector. Last, the detector may be used for evaluating the target dataset.} \label{fig:intro}
\end{figure}

Weakly/semi-supervised methods \cite{hu2017wordsup,qin2019curved,rong2017weakly,tian2017wetext} are usually utilized together to reduce the need for complex annotations. However, most semi-supervised methods rely heavily on the annotations of the target domain. Although weakly supervised methods can reduce the cost in the labeling process, they still need a large number of annotated samples. Although data generation methods such as \cite{gupta2016synthetic,liao2019synthtext3d,zhan2018verisimilar,zhan2019spatial} use prior knowledge to automatically render texts in non-text images, but the data they generate are not ``realistic'' enough and the background images are limited. There is a big performance gap between the methods using generated data and those using real data. Recently, the domain adaption method \cite{zhan2019ga} tries to improve the performance on the target domain with informative source domain samples.

Compared with above methods, self-training (Fig.~\ref{fig:intro}) is an alternative to solve the cross-domain problem. Instead of generating ``real" data, a large number of real images and videos can be utilized to extract useful information to re-train the model from source domain. In this way, the performance and generalization ability of the detector can be improved without any manual annotation.
%
The cross-domain of general object detection has received lots of attention, but the cross-domain of text detection has not been studied much. Adaptations between virtual/real domains, multi-language domains, document/scene text domains, etc. are ubiquitous but cross-domain performance degradation is drastic.

\begin{figure*}[t]
\centering
\includegraphics[width=2\columnwidth]{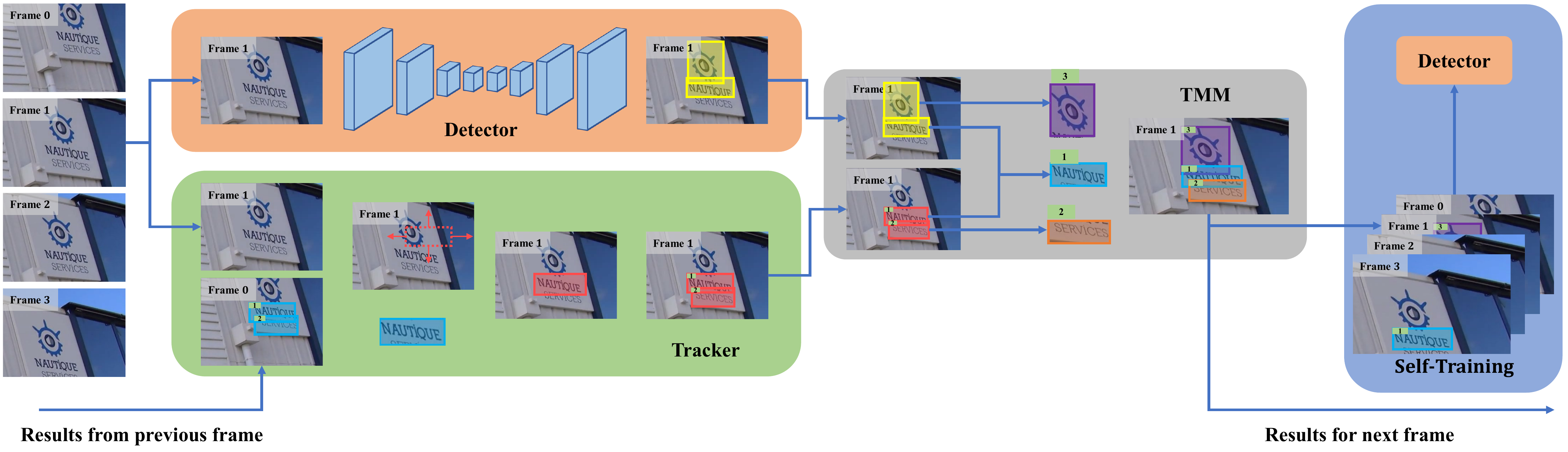}
\caption{Illustration of the self-training framework. Boxes marked by \textbf{yellow} represent the detection results. Boxes marked by \textbf{red} represent the tracking results. Boxes marked by \textbf{blue, purple (hard negative) and orange (hard positive)} are trajectory results. Each frame of the video will be fed into the detector, tracker, and TMM to get the pseudo-labels. Pseudo-labels together with video frames will be used for re-training of the detector. Detector here is a deep CNN module. Tracker uses the location (\textbf{red dash-box}) of the entity and its appearance feature (\textbf{blue cropped box}) to search the corresponding entity in the current frame. TMM is responsible for updating the trajectory. 
(Zoom-in view for more details )} 
\label{fig:achi}
\end{figure*}
Inspired by \cite{roychowdhury2019automatic} and \cite{yang2017tracking}, we propose a new self-training framework for text detection. This framework tries to automatically find hard examples from massive unannotated images and videos, and it consists of a detector to provide initial detection results and a tracker to provide tracking results. Since texts tend to appear densely and have relatively blurred boundary partitions, a new Text Mining Module (TMM) is proposed to fuse and filter the detection and tracking results. These three modules work together to accurately find hard examples and reduce the noise rate for the self-training procedure. For some applications where only images should be processed, we design an image-to-video generation method to generate videos suitable for our framework.
\begin{itemize}
\item[]  The contributions are as follows:   
 \item We design a self-training framework for domain adaptive scene text detection using unannotated videos or images, which achieve comparable or even superior results over the state-of-the-art methods. This is the first work to address the problem of domain adaptation in scene text detection with self-training. 
 \item A novel text mining module is proposed to effectively mine hard examples of less noise rate based on the fusion of detection and tracking results, which is specially designed for hard text example mining.

 \item We have designed an image-to-video generation method (Gen-Loop) for the tasks that videos are unavailable and only images can be used, and the generation method can also avoid processing non-text video frames that can speed up the mining process.
\end{itemize}


\section{Related Work}
Methods of using less annotated data to train scene text detector can be roughly divided into three categories. The first is weakly/semi-supervised method \cite{hu2017wordsup,qin2019curved,rong2017weakly,tian2017wetext}. These works use weak annotations including bounding box or text/non-text label for character detection, activation map generation or mask generation. Both weakly and semi-supervised methods still rely on data annotation more or less. The second is data generation methods \cite{gupta2016synthetic,liao2019synthtext3d,zhan2018verisimilar}. These methods use prior knowledge, statistical information and 3-D depth maps to generate synthetic text images. There is a huge gap between the performance of models trained with synthetic data and those with real data. The third is the domain adaption method. \cite{zhan2019ga} designs an innovative geometry-aware domain adaptation network that learns and models domain shifts in appearance and geometry spaces simultaneously. The bounding box annotation is required to crop the text regions.

The self-training method uses existing detectors to generate pseudo-labels on unlabeled data sets, and selects a subset with high confidence of the pseudo-labels for re-training. \cite{rosenberg2005semi} uses the detection results from a pre-trained object detector on unlabeled data as pseudo-labels and then trains on a subset of this noisy labeled data in an incremental re-training procedure. \cite{jin2017end} uses tracking in videos to gather hard examples, then re-trains the detector using this extra data to improve detection on still images. \cite{roychowdhury2019automatic} uses self-training to seek automatic adaptation to a new target domain. Their methods aim to automatically obtain hard examples of general objects, faces, or pedestrians but fail for texts because texts usually appear densely and have relatively blurred boundary partitions. 

Several works are concentrating on video text detection. \cite{yang2017tracking} first generates trajectories by multi-strategy tracking techniques, and then links the tracking trajectories by all detection, recognition and prediction information using a tracking network (graph). \cite{cheng2019efficient} trains a spatial-temporal text detector for localizing text regions, and each tracked text stream is recognized one-time with a text region quality scoring mechanism. They aim to detect texts on videos using sequence information, while our work aims to improve text detection on still images using video data of the target domain.

\section{Proposed method}

This section is organized as follows. First, we give an overview of the framework. Then, we illustrate the effectiveness of the TMM on reducing noise rate. Next, we explain the design of the image-to-video generation method. Finally, the loss function is given.

\subsection{Framework Overview}

\begin{figure}[t]
\centering
\includegraphics[width=1\columnwidth]{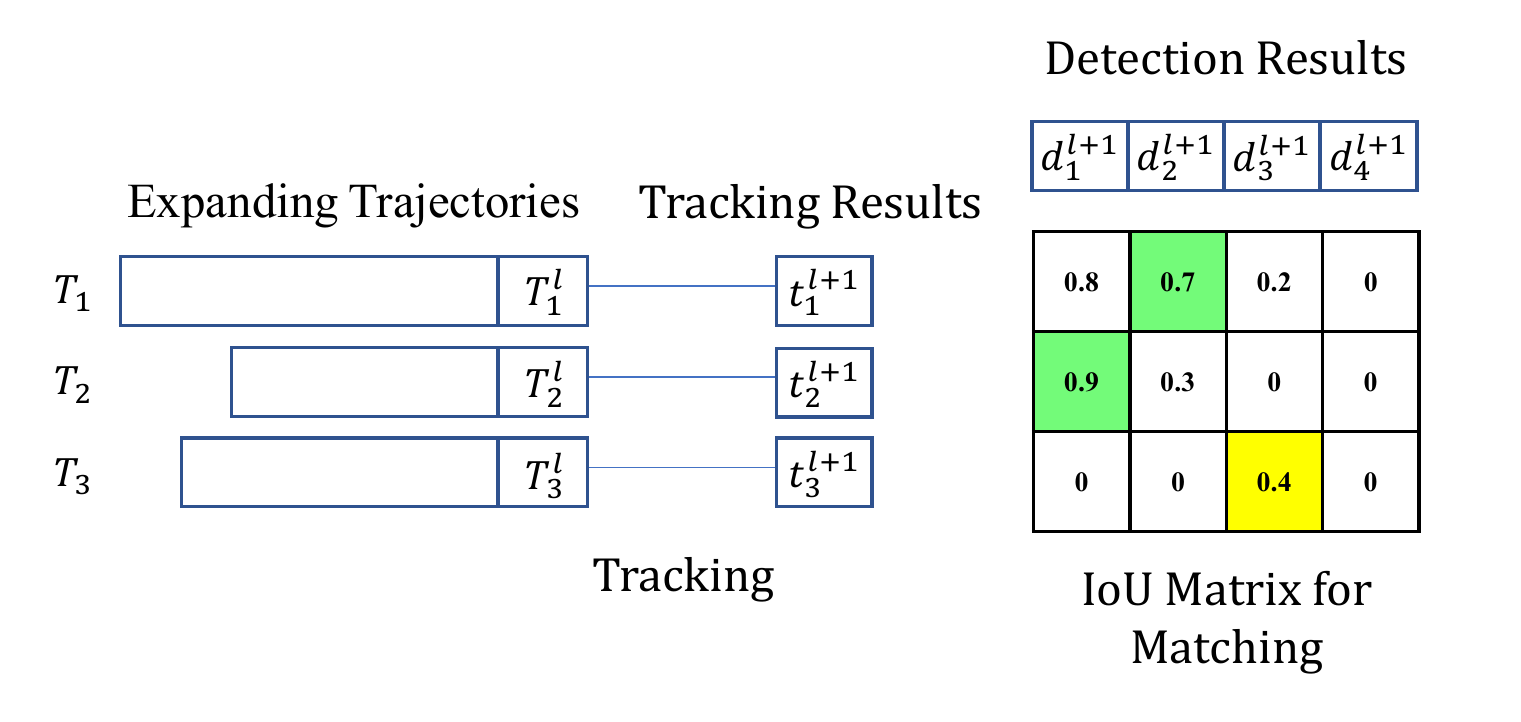}
\caption{Illustration of the relationship among trajectories, tracking results and detection results. Green cells represent the successful matching, yellow cell represents a weak match that should be ignored. } \label{fig:match}
\end{figure}
As illustrated in Fig.~\ref{fig:achi}, the framework is composed of three key modules: the detection module, the tracking module, and the TMM. We train an initial detector on the training dataset of the source domain, and prepare a large amount of video data of the target domain. 
The video data are processed by the detector and the tracker, and TMM can automatically mine the hard examples with pseudo-labels using the detection and the tracking results. If and only if there are hard positives ${H\!P}_{A}$ or hard negatives ${H\!N}_{A}$ in image A, A is added to the pseudo dataset. The corresponding pseudo-labels $\tilde{L}_A$ for A are calculated by
\begin{equation}
\tilde{L}_A=D_A\backslash H\!N_A\cup {H\!P}_A
\end{equation}
where $D_A$ is the detection results in image A and $D_A\backslash H\!N_A$ represents the set of elements in $D_A$ but not in $H\!N_A$. 
Then the pseudo dataset can be used for the re-training of the detector.

Any detector and tracker can be used in the framework. We choose Mask R-CNN \cite{he2017mask} as the baseline detector in that it can achieve state-of-the-art performance on most benchmarks. Considering the simplicity and processing speed, we choose the template matching algorithm as our tracker. The tracker is used to generate tracking results in the current frame. Trajectories are generated by fusing the detection results and the tracking results. Each trajectory is used to describe the locations of a specific instance in the video. The location of the text instance is represented by a bounding box and an instance segmentation mask.

TMM is responsible for updating the trajectory. It is designed to handle three different tasks. The first task is to select a suitable detection result or a tracking result as the final trajectory result. The second task is to predict an estimated segmentation mask for the trajectory result. And the third task is to perform hard example mining. In a trajectory, if a single tracking result appears both following and followed by several consecutive detection results,  the tracking result is what we meant to be a hard positive. And if some trajectories are too short or have few detection results, we consider detection results in such trajectories as hard negatives.

\subsection{Text Mining Module}

TMM is the key component of the framework, which is designed to fuse the detection results and the tracking results. In general, the maximum intersection-over-union (IoU) is the evaluation metric for judging whether a detection result or a tracking result matches to a trajectory. As shown in Fig.~\ref{fig:match}, the last item $T_j^l$ in a trajectory $T_j$ is the trajectory result of instance $j$ in the frame $l$. In the frame $l+1$, $T_j^l$ will have a tracking result, denoted as $t_j^{l+1}$. Once $t_j^{l+1}$ is generated, it is temporarily joined into $T_j$. For $T_j$, $k$ is the matching index of all detection results, which is calculated by
\begin{equation}
k=
\begin{cases}
\arg\max\limits_i(F(i,j)),& if\; \max(F(i,j))>\theta\\
N\!one,&otherwise
\end{cases}
\end{equation}
\begin{equation}
\label{eq_m}
F(i,j) = max(IoU(d_i^{l+1},t_j^{l+1}), IoU(d_i^{l+1},T_j^l))
\end{equation}
where $\theta$ is the threshold for IoU matching, $d_i^{l+1}$ represents the $i_{th}$ detection result in the frame $l+1$, ``$N\!one$'' means there is no matching result for $T_j$. If $k\ne N\!one$ then $t_j^{l+1}$ will be replaced by $d_k^{l+1}$. So either a detection result or a tracking result will be joined into $T_j$. In this way the detection and tracking information can be fused to obtain a more accurate trajectory. If the detection result does not match any trajectory, it will be initialized as a new one.

\begin{figure}[t]
\centering
\includegraphics[width=1\columnwidth]{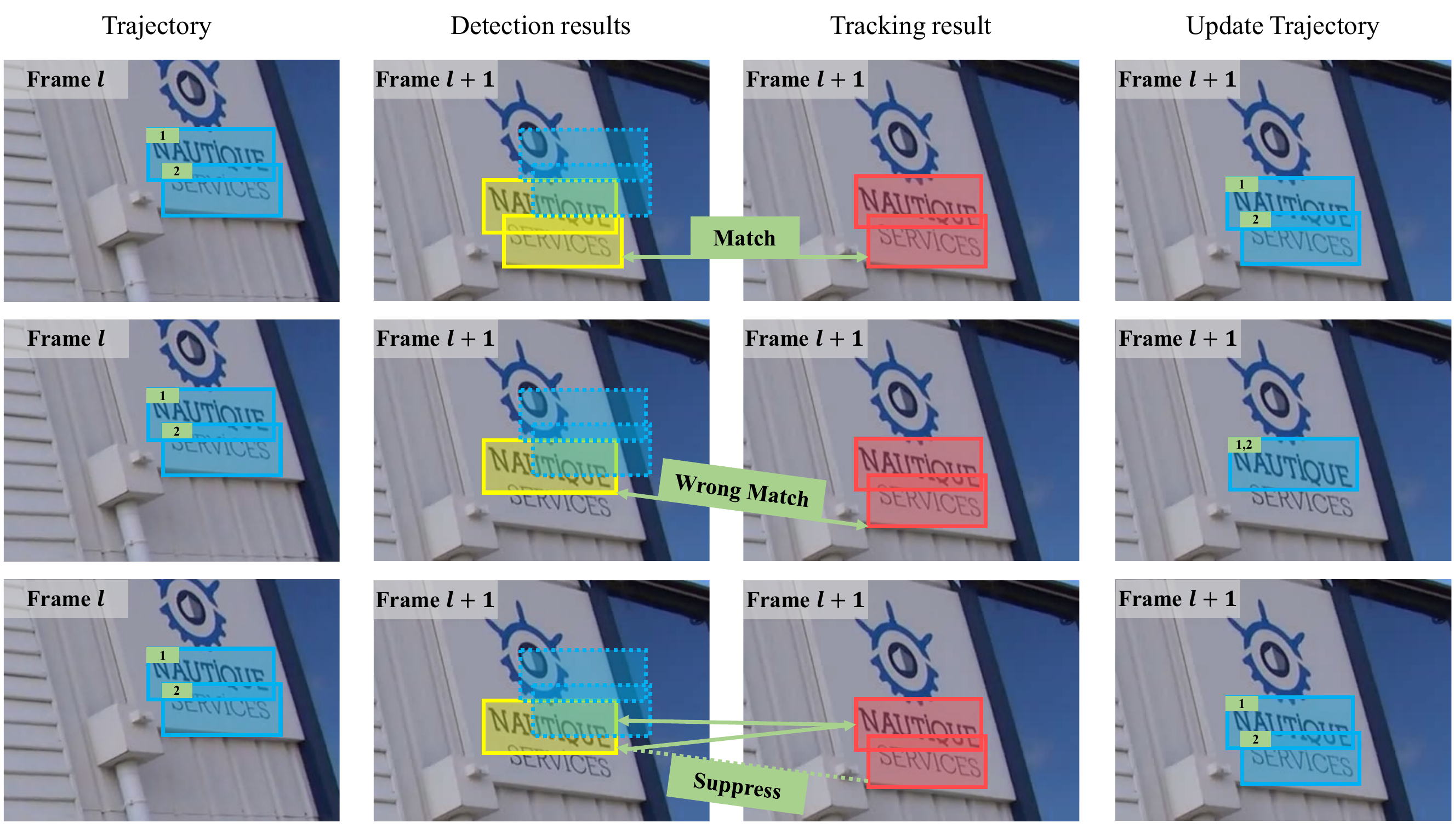}
\caption{Visualization of trajectory results. The meaning of different colors can refer to Fig.~\ref{fig:achi}. Row 1: a correct trajectory updating. Row 2: a false trajectory updating with general matching rules when some detection results are missing. Row 3: trajectory updating results with TMM when some detection results are missing.}
\label{fig:diff}
\end{figure}

The matching method mentioned above does not always work well. For the reason that the videos are from the target domain, we are prone to encounter the missing of detection results or tracking results. When a trajectory is expanded, it follows the principle of “first come, first served”. That is, if a detection result is matched by the current trajectory, it can not be matched by any other trajectory $T_o$ even if the detection result has a much larger IoU with $T_o$. This simple method can work well when targets are sparse, but text instance tends to appear densely and has relatively blurred boundary partitions, leading to a more complicated situation, and it will cause confusing tracking results. As shown in row 2 of Fig.~\ref{fig:diff}, two trajectories will focus on the same instance “NAUTIOUE”.
To solve this problem, TMM needs to consider not only which trajectory each detection result should match, but also which detection result each trajectory should match. In other words, previous greedy match problem should be converted to a search problem. We use a Matrix $M_{IoU}$ to save all results of Equ.~\ref{eq_m}. The dimension of $M_{IoU}$ is the number of detection results $N_d$ times the number of trajectories $N_j$. For each detection result $d_i$, we find out the trajectory $T_j$ which has the largest IoU with $d_i$, and then search in $M_{IoU}$ to find out whether $d_i$ has the largest IoU with $T_j$. A successful matching should satisfy both Equ.~\ref{eq_match1} and Equ.~\ref{eq_match2}, where ``=='' represents equal function. If $d_i$ does not match $T_j$, $M_{IoU}(i,j)$ is suppressed to 0, and a new search is performed again for $d_i$ until the matching result is found. (See row 3 of Fig.~\ref{fig:diff}) Once we find the matching result, the tracking result in $T_j$ will be replaced by $d_i$. If no matching result is found, the $d_i$ is initialized as a new trajectory.

\begin{equation}
\label{eq_match1}
i==\arg\max\limits_p(M_{IoU}(p,j)) 
\end{equation}
\begin{equation}
\label{eq_match2}
j==\arg\max\limits_p(M_{IoU}(i,p))
\end{equation}
\begin{figure}[t] 
\centering
\includegraphics[width=1\columnwidth]{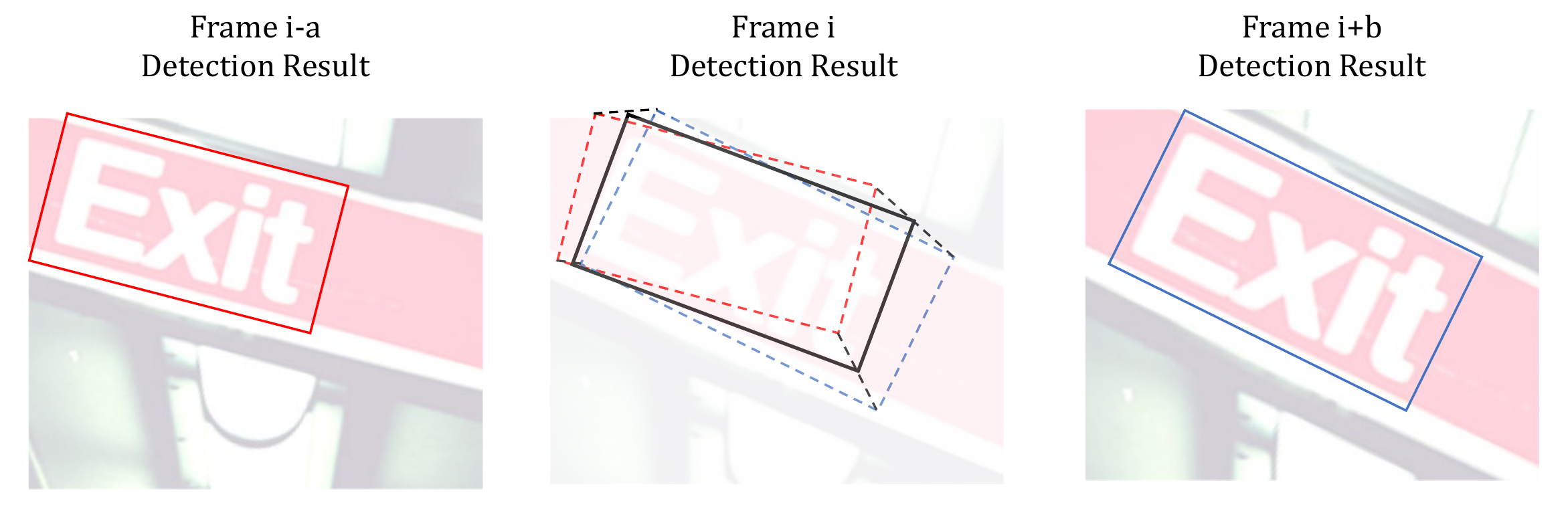}
\caption{Visualization of the mask estimation. Frame i-a and frame i+b are the closest frames that have detection results of instance “Exit” to frame i. The red rectangle and blue rectangle are the minimum bounding rectangle of the segmentation mask. The black polygon is the estimated segmentation mask of the tracking result.} 
\label{fig:est}
\end{figure}

Scene our detector is Mask R-CNN, we need the instance segmentation results of it to get a more accurate representation of texts, but most trackers can't produce the segmentation mask. TMM should predict an estimated segmentation mask for a tracking result in a trajectory. 
The video is continuous and the time interval between two frames is so short that we can assume the perspective transformation of the text area is uniform. We find two closest detection results forward and backward, and the minimum bounding rectangles of the segmentation masks of the detection results are obtained. As shown in Fig.~\ref{fig:est}, the corresponding points of detection results are the inputs of linear interpolation function to get a polygon as the estimated segmentation mask for the tracking result.

\subsection{When No Video Available}

The above self-training framework can be applied in any target domain as long as videos can be obtained. However, in some applications only images should be processed but no videos. Moreover, a large number of video frames do not contain text regions, which will be meaningless and time-consuming to process. If we could collect images containing scene texts from the target domain, can we make it useful?

A straightforward method (Base) can use images directly for self-training, which means we get pseudo-labels only by detectors for the images and use them for re-training. Otherwise, we can use some sophisticated data enhancement techniques (Gen-Straight) to generate synthetic videos. For an input image, we can randomly generate a rotation angle $\theta$, a scaling factor $\delta$, and a transformation center $c$. We can use the affine transformation matrix based on these parameters to generate the ending frame of the video. Once the images of the starting frame $i$ and the ending frame $j$ are obtained, a video of length $t$ can be generated by interpolation, with which hard examples can be mine using above framework. For a fair comparison, we will perform the same transformation on the pseudo dataset generated by Base, which is called Base-Trans.

\begin{figure}[t]
\centering
\includegraphics[width=1\columnwidth]{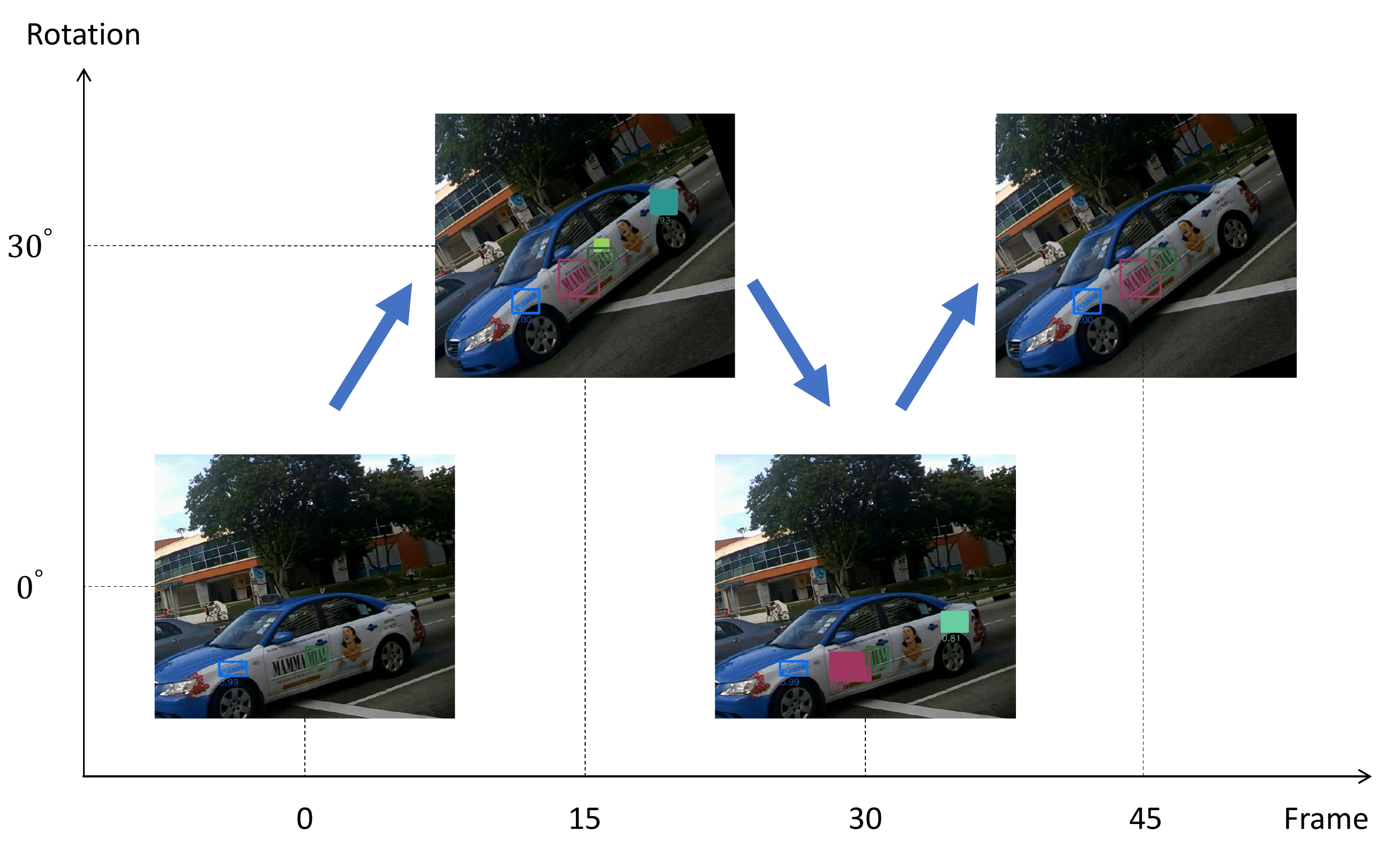}
\caption{Visualization of trajectories on video generated by the Gen-Loop method. Boxes marked by the same color are from the same trajectory. Hollow boxes with masks are the detection results, solid boxes are the tracking results which should be considered as hard positives.} 
\label{fig:img2vid}
\end{figure}

Unfortunately, we can not get hard examples effectively if we use the naive method directly. For the detection task, the frames generated by Gen-Straight is either from difficult to easy or from easy to difficult. Once the detection result of a frame is lost, it is foreseeable that all frames following or followed by it will be lost. More importantly, it is impossible to mine hard examples from the starting and the ending frame because they have no adjacent frame either before or after them, while the hard examples in the starting frame are the most valuable information for us. Consequently, we design a loopback (Gen-Loop) to trace from frame $i$ to $j$, then from frame $j$ to $i$, and finally from frame $i$ to $j$ again. As illustrated in Fig.~\ref{fig:img2vid}, this ensures that each image in the sequence is visited at least twice, and at the same time all images have adjacent frames on both sides. The length $t$ of the video is limited to 50 to ensure that the time interval between two adjacent visits is not too long. This new generation pattern is very effective for the self-training framework. Moreover, because each generated frame is duplicated three times, only 1/3 of the normal workload is required for the detection module.

\subsection{Balance Loss}
Because of the limitations of detector performance, tracker performance and video quality, it is difficult to ensure that the data we mine are noise-free. Effective noise suppression method is very critical. \cite{roychowdhury2019automatic} shows that the loss function can be an extra way to suppress noise in self-training methods. Here we use a refined binary cross-entropy in our detector for $L_{cls}$ in Mask R-CNN \cite{he2017mask}.

\begin{equation}
\tilde{y_i}=
\begin{cases}
y_i, &if\; X_i\; has\; annotation\\
1, &elif\; X_i \in H\!P\\
s_t, &else
\end{cases}
\end{equation}
\begin{equation}
\label{eq_loss}
L_{cls}(\tilde{y_i},p_i)=-[\tilde{y_i}\log(p_i)+(1-\tilde{y_i})\log(1-p_i)]
\end{equation}
where $X_i$ is one of the samples from both the annotated and pseudo dataset, $y_i$ is the corresponding annotation of $X_i$ if $X_i$ has annotation. $H\!P$ means the datasets of hard positives. $s_t$ is the detection or tracking score of the pseudo label. Balance loss can give different samples different labels. In other words, it can increase the training weight of hard examples, and reduce the training weight of low confidence samples.


\begin{table*}[t]
\begin{center}
\caption{Ablation study on IC15 and 15VID test set with different training datasets and different modules. The source datasets are VISDT and IC15, and the target datasets are IC15 and 15VID.} \label{tab:ablation}
\begin{tabular}{|c|c|c|c|c|c|c|c|c|c|}
\hline
\multirow{2}{*}{DATASET}&\multirow{2}{*}{15VID} & \multirow{2}{*}{TMM} & \multirow{2}{*}{Balance Loss}  & \multicolumn{3}{c|}{IC15}& \multicolumn{3}{c|}{15VID}\\
   \cline{5-10}
&           &           &           & Precision & Recall   & F-measure & Precision & Recall & F-measure  \\
\hline
\multirow{4}{*}{VISDT}
& \ding{55} & \ding{55} & \ding{55} & 54.8      & 57.7      & 56.2      & 49.8      & 48.0      & 48.9\\
& \ding{52} & \ding{55} & \ding{55} & \bf{67.7} & 52.1      & 58.9      & 55.7      & 49.3      & 52.3\\
& \ding{52} & \ding{52} & \ding{55} & 62.3      & 61.6      & 61.9      & 56.3      & \bf{54.0} & 55.1\\
& \ding{52} & \ding{52} & \ding{52} & 64.3      & \bf{61.7} & \bf{63.0} & \bf{60.9} & 53.6      & \bf{57.0}\\
\hline                
\multirow{4}{*}{IC15} 
& \ding{55} & \ding{55} & \ding{55} & 83.0      & 80.4      & 81.7      & 63.5      & 60.2      & 61.8\\
& \ding{52} & \ding{55} & \ding{55} & 83.9      & 81.1      & 82.5      & 64.1      & 60.7      & 62.4\\
& \ding{52} & \ding{52} & \ding{55} & 85.4      & \bf{81.3} & 83.3      & 65.4      & 60.5      & 62.9\\
& \ding{52} & \ding{52} & \ding{52} & \bf{87.7} & 80.3      & \bf{83.8} & \bf{65.4} & \bf{61.9} & \bf{63.6}\\
\hline

\end{tabular}
\end{center}
\end{table*}

\begin{table}[t]
\begin{center}
\caption{Results of iteration on IC15 test set. Iter 0 means the result of initial detector trained by VISDT, iter n means the result of the n-th iteration of re-training.} 
\label{tab:iter}
\begin{tabular}{|c|c|c|c|c|}
\hline
Iteration & Precision & Recall & F-measure\\
\hline%
iter 0 & 54.8      & 57.7      & 56.2 \\
iter 1 & 64.3      & 61.7      & 63.0 \\
iter 2 & \bf{70.5} & 64.7      & \bf{67.5} \\
iter 3 & 67.3      & \bf{66.4} & 66.8 \\
\hline
\end{tabular}
\end{center}
\end{table}

\begin{table}[t]
\begin{center}
\caption{Comparison with Data Augmentation on IC15 test set. Aug represents data augmentation.} \label{tab:diff-det}
\begin{tabular}{|l|c|c|c|}
\hline
Method              & Precision & Recall    & F-measure     \\
\hline
Mask R-CNN          & 83.0      & 80.4      & 81.7 \\
Mask R-CNN+Ours     & 87.7      & 80.3      & 83.8 \\
Mask R-CNN+Aug      & 86.1      & \bf{84.1} & 85.1 \\
Mask R-CNN+Aug+Ours & \bf{89.8} & 82.5      & \bf{86.0} \\
\hline
\end{tabular}
\end{center}
\end{table}

\section{Experiments}
We evaluate our approach on three standard benchmarks: ICDAR2015, ICDAR2017 MLT, MSRA-TD500, and compare with other state-of-the-art methods.

\subsection{Datasets}
The datasets used for the experiments in this paper are briefly introduced below:

\textbf{ICDAR2015 Text in Video (15VID)}
\cite{karatzas2015icdar} includes a training set of 25 videos (13450 frames in total) and a test set of 24 videos (14374 frames in total), 1k frames with annotations are randomly selected from it to form the test set for domain adaptation. 25 videos without annotations are used as the input of the framework.

\textbf{ICDAR2015 (IC15)}
\cite{karatzas2015icdar} is a dataset proposed for incidental scene text detection. There are 1000 training images and 500 test images with annotations labeled as word-level quadrangles. 

\textbf{Verisimilar Image Synthesis Dataset (VISD)}
is a synthetic dataset proposed in \cite{zhan2018verisimilar}. It contains 10k images synthesized with 10k background images. To facilitate comparison, we randomly select 1k images from the dataset and mark them as VISDT.

\textbf{ICDAR2017 MLT (MLT17)}
\cite{nayef2017icdar2017} consists of 7200 training images, 1800 validation images, and 9000 test images. Image annotations are labeled as word-level quadrangles. We use only the training set to train our model.

\textbf{MSRA-TD500 (TD500)}
\cite{yao2012detecting} consists of 500 natural images (300 for training, 200 for test), which are taken from indoor and outdoor scenes using a pocket camera.
All the images in the dataset are annotated in the text-line level.
\subsection{Implementation Details}
We set hyper-parameters mainly following MMDetection\footnote{https://github.com/open-mmlab/mmdetection}. Our base-model is ResNet-50 and pre-trained on ImageNet. We use SGD as the optimizer with batch size 4 for training. On IC15 and MLT17, we train our model for 12 epochs with the learning rate of 0.005, and the learning rate is reduced by 10 times at the 8th epoch. On TD500, we train our model for 60 epochs with the same initial learning rate, and the learning rate is reduced by 10 times at the 40th epoch. If the pseudo dataset B is generated by the detector trained by dataset A, we will use both B and A for the re-training. The aspect ratios of anchors are set to 1/5, 1/2, 1, 2, 5 for all experiments. We use minAreaRect in OpenCV to obtain the quadrilateral bounding boxes of text instances segmentation as the final detection result. 

For data augmentation, a random horizon flip is applied with a probability of 0.5. Images are rotated in range $[0^{\circ},90^{\circ}]$ randomly. Random crop and resize and multi-scale training are also applied. When applying augmentation, 4x learning rate scheduler is used.
\subsection{Ablation Study}
To verify the effectiveness of our approach, we do a series of ablation studies on different training sets, different test sets, and different modules. First, we train a basic detector on the synthetic dataset VISDT and IC15 as the initial detectors. For different detectors we use the training set of 15VID as the input of our framework to get the corresponding pseudo-labels, after hard example mining, we will randomly extract 1000 images among all hard examples. Together with the generated pseudo labels, they make up the pseudo dataset. Different initial detectors are used to test the effectiveness of our framework under different situations. Results on IC15 are used to measure detection performance and results on 15VID are used to test the performance of domain adaptation.

\begin{figure}[t]
\centering
\includegraphics[width=1\columnwidth]{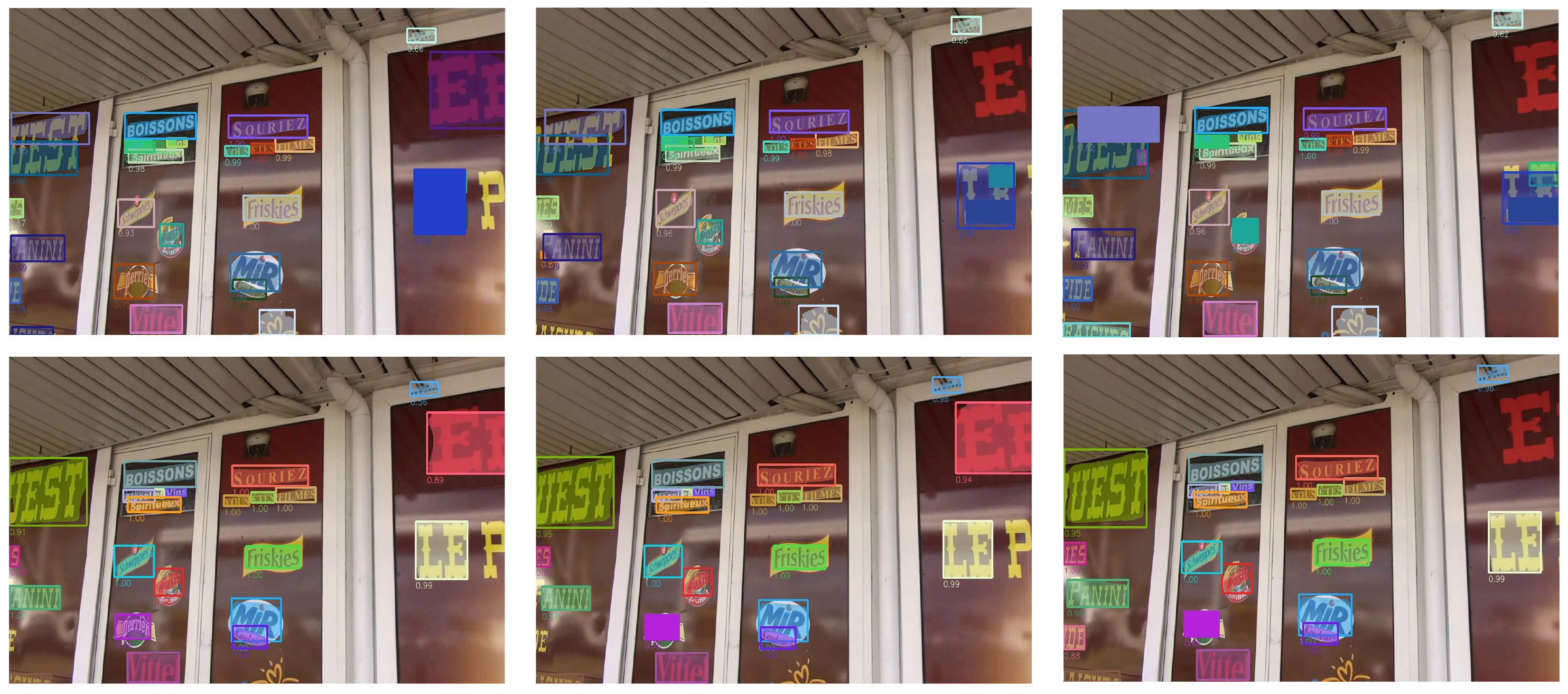}
\caption{Illustration of results generated by different detectors. Boxes or mask marked by the same color represents the locations of the same entity. Solid rectangle boxes represent hard positives. Row 1: results of the initial detector trained by VISDT. Row 2: results of detectors after self-training. } \label{fig:quality}
\end{figure}

On this basis, we design some experiments to verify the performance of all methods mentioned in Section 3. Table~\ref{tab:ablation} summarizes the results of our framework with different settings on IC15. The experiments with dataset VISDT are mainly used to verify the domain adaptation performance of our framework, when no information of the target domain is used, only 56.2\% and 48.9\% in f-measure can be achieved on test set. Using the video from 15VID can make about 2.7\% and 3.4\% improvements on F-measure. While using TMM to improve the quantity and quality of hard examples and to generate more pseudo annotations that greatly improve recall, further improvements of 3\% and 2.8\% in f-measure can be achieved. Last, given different weights to detection results and hard examples can also bring improvements to the model. The experiments with dataset IC15 proves that our framework does not sacrifice the performance on the source domain to perform domain adaptation. It can be seen that even with better initial detectors, the growth pattern under different settings has not changed. Retrained models can outperform initial models 2.1\% and 1.8\% in f-measure. These experiments show that our framework can not only be used in domain adaptation tasks but also be used to improve the performance of detectors. Fig.~\ref{fig:quality} shows qualitative results of the initial detector and the detector after self-training.
\begin{table}[t]
\begin{center}
\caption{Results on MLT17-KOR.} \label{tab:multilan}
\begin{tabular}{|l|c|c|c|}
\hline
Method        & Precision & Recall & F-measure     \\
\hline
17NK          & 65.1 & 49.7 & 56.4 \\
17NK+PSD\_NK  & 67.7 & 64.5 & 66.1 \\
MLT17         & 80.9 & 70.6 & 75.4 \\
MLT17+PSD\_17 & \bf{82.8} & \bf{71.3} & \bf{76.6} \\
\hline
\end{tabular}
\end{center}
\end{table}

\subsection{Results of Iteration}

We design an additional experiment to explore the impact of iteration and the upper bound of it. After being re-trained with our framework, a detector can be regarded as the initial detector to mine new hard examples on the same video datasets. The results of Table~\ref{tab:ablation} shows that great improvement can be achieved through self-training. The performance gap will lead to different hard examples on the same video dataset. Through iteration, we will get more annotated data and richer background information.
Table~\ref{tab:iter} shows the impact of iteration. We can see that the 2nd iteration can still achieve a great improvement, but the performance of the third iteration begins to decline. Since we can not eliminate all the noise when mining hard examples, the accumulation of noise reaches the upper limit and leads to performance reduction.

\subsection{Comparison with Data Augmentation}

\begin{table}[t]
\begin{center}
\caption{Results with different video-lacking strategies on IC15 test sets.} \label{tab:genvid}
\begin{tabular}{|l|c|c|c|}
\hline
Method             & Precision & Recall    & F-measure     \\
\hline
None               & 54.8      & 57.7      & 56.2 \\
Base               & 62.7      & 58.9      & 60.7 \\
Base-Trans         & 64.1      & 59.1      & 61.5 \\
Gen-Straight\;\;\; & 64.4      & 60.0      & 62.1 \\
Gen-Loop           & \bf{66.9} & \bf{64.7} & \bf{65.8} \\
\hline
\end{tabular}
\end{center}
\end{table}

Our method attempts to find hard examples from videos and use them for re-training, which seems similar to data augmentation, but they are essentially different. To verify this, we design some experiments with and without data augmentation, results are in Table~\ref{tab:diff-det}. It can be seen that even with data augmentation, our framework can still achieve a nearly 1\% improvement on f-measure. Both of them can make the target pattern richer, but our approach can get more abundant backgrounds, and can focus on samples that the detector can't handle.

\subsection{Results on Multilingual Environment}
\begin{table}[t]
\begin{center}
\caption{Comparison with cross domain and data generation methods on the ICDAR2015 test images. ``Target'' indicates the training set of the target dataset. Top: results without target dataset. Bottom: result with target dataset.} \label{tab:generating}
\small
\begin{tabular}{|l|c|c|c|}
\hline
Method                     & P & R & F     \\
\hline
Synthtext3d\cite{liao2019synthtext3d} & 64.5      & 56.7      & 60.3 \\
GA-DAN\cite{zhan2019ga}[AD]           & 69.9      & 59.6      & 64.4 \\
GA-DAN\cite{zhan2019ga}[10-AD]        & 67.3      & \bf{71.6} & \bf{69.4} \\
VISDT                                 & 54.8      & 57.7      & 56.2 \\
VISDT+15VID                           & 64.3      & 61.7      & 63.0  \\
VISDT+15VID-2                         & \bf{70.5} & 64.7      & 67.5 \\
\hline
Synthtext3d\cite{liao2019synthtext3d} & 86.6      & 79.2      & 82.7  \\
GA-DAN\cite{zhan2019ga}[AD]           & 83.7      & 79.2      & 81.4 \\
GA-DAN\cite{zhan2019ga}[10-AD]        & 85.6      & 81.6      & 83.5  \\
Target                                & 80.3      & 81.7      & 81.0  \\
VISDT$\rightarrow$Target              & 83.0      & \bf{82.2} & 82.6   \\
15VID$\rightarrow$Target              & \bf{86.9} & 81.7      & \bf{84.2}  \\
\hline

\end{tabular}
\end{center}
\end{table}

The experiments above show that the detector can be greatly improved by self-learning without using any manually annotated data, which is valuable for practical applications. Here we consider another case where a well-trained multilingual detector is used to detect texts of a new language. A straightforward method is to collect annotated images of the new language, and then fine-tunes the detector. There are too many languages all over the world, which leads to difficulties in annotation. As an illustration, we select Korean as the target language. We annotate all Korean data in MLT17 as NOT-CARE and get a subset represented as 17NK and all Korean images from the validation set are used as the test subset MLT17-KOR. We download some videos of Korean language from YouTube as the input of our framework. Table~\ref{tab:multilan} shows the experimental results. PSD\_NK and PSD\_17 are the data mined in video by detectors trained on 17NK and MLT17 respectively. Although the multilingual model does not contain any information about the Korean language, the model still has detection capabilities due to the similarity of texts. Under such a situation, our framework can improve itself by self-training.

\subsection{When No Video Available}
When we can only get images, we design a new video generation mode (GEN-LOOP) for our framework. Table~\ref{tab:genvid} records our experimental results. “None” indicates the result of the initial detector trained by VISDT. The remaining descriptions are shown in section 3.3. It can be seen that the result of using images for re-training is relatively low, because we can't use the continuity of the video to reduce the noise during re-training. Simple data augmentation can increase by 0.8\% on f-measure, but it does not solve the underlying problem. The following two methods both use images to generate video, but Gen-Straight is 3\% lower than Gen-Loop. This is because this generating pattern conflicts with the method of hard example mining. So we can not find samples which are valuable enough for re-training.

\begin{table}[t]
\begin{center}
\caption{Experimental results on ICDAR 2015 dataset. ``Ext.'' indicates external data with annotations.  For fair comparison, this table only listed the single scale results without recognition supervision.} \label{tab:compare_15}
\begin{tabular}{|l|c|c|c|c|}
\hline
Method                                & Ext.      & P & R & F     \\
\hline
PAN\cite{Wang2019efficient}           & -         & 82.9      & 77.8      & 80.3 \\
PSENet\cite{Wang2019shape}            & -         & 81.5      & 79.7      & 80.6 \\
Synthtext3d\cite{liao2019synthtext3d} & -         & 86.6      & 79.2      & 82.7 \\
GA-DAN\cite{zhan2019ga}               & -         & 85.6      & 81.6      & 83.5 \\
Ours[IC15]                            & -         & 80.3      & 81.7      & 81.0 \\
Ours[15VID$+$IC15]                    & -         & 85.4      & 81.3      & 83.3 \\
Ours[15VID$\rightarrow$IC15]          & -         & \bf{86.9} & \bf{81.7} & \bf{84.2} \\
\hline                                            
MSR\cite{Xue2019msr}                  & \ding{52} & 86.6      & 78.4      & 82.3 \\
Mask TextSpotter\cite{yao2018mask}    & \ding{52} & 85.8      & 81.2      & 83.4 \\
FOTS\cite{liu2018fots}                & \ding{52} & 88.8      & 82.0      & 85.3 \\
PSENet-1s\cite{Wang2019shape}         & \ding{52} & 86.9      & 84.5      & 85.7 \\
BDN \cite{liu2019omnidirectional}     & \ding{52} & 89.4      & 83.8      & 86.5 \\
SPCNet \cite{xie2019scene}            & \ding{52} & 88.7      & 85.8      & 87.2 \\
LOMO \cite{zhang2019look}             & \ding{52} & \bf{91.3} & 83.5      & 87.2 \\
GNNets \cite{xu2019geometry}          & \ding{52} & 90.4      & \bf{86.7} & \bf{88.5} \\
Ours[IC15]                            & \ding{52} & 88.0      & 83.9      & 85.9 \\
Ours[15VID$\rightarrow$IC15]          & \ding{52} & 88.3      & 85.7      & 87.0 \\
Ours[15VID$\rightarrow$IC15]+AUG      & \ding{52} & 91.2      & 85.4      & 88.2 \\
\hline
\end{tabular}
\end{center}
\end{table}  

\begin{table}[t]
\begin{center}
\caption{Experimental results on MSRA-TD500 benchmark with single scale testing. ``Ext.'' indicates external data with annotations. } \label{tab:compare_td}
\begin{tabular}{|l|c|c|c|c|}
\hline
Method                            & Ext.      & P & R & F     \\
\hline
GA-DAN\cite{zhan2019ga}           & -         & 80.5      & 71.1      & 75.5 \\
EAST\cite{zhou2017east}           & -         & \bf{87.3} & 67.4      & 76.1 \\
He etc.\cite{he2018multi}         & -         & 85.0      & 70.0      & 76.7 \\
PAN\cite{Wang2019efficient}       & -         & 80.7      & 77.3      & 78.9 \\
Ours[TD500]                       & -         & 71.4      & 68.7      & 70.0 \\
Ours[15VID$\rightarrow$TD500]     & -         & 79.5      & \bf{78.7} & \bf{79.1} \\
\hline                                         
DSRN\cite{Wang2019dsrn}           & \ding{52} & 87.6      & 71.2      & 78.5 \\
Text Field\cite{xu2019textfield}  & \ding{52} & 87.4      & 75.9      & 81.3 \\
MSR\cite{Xue2019msr}              & \ding{52} & 87.4      & 76.7      & 81.7 \\
CRAFT\cite{baek2019character}     & \ding{52} & 88.2      & 78.2      & 82.9 \\
MCN\cite{liu2018learning}         & \ding{52} & 88.0      & 79.0      & 83.0 \\
PAN\cite{Wang2019efficient}       & \ding{52} & 84.4      & 83.8      & 84.1 \\
BDN \cite{liu2019omnidirectional} & \ding{52} & \bf{89.6} & 80.5      & 84.8 \\
Ours[TD500]                       & \ding{52} & 83.3      & 80.8      & 82.0 \\
Ours[15VID$\rightarrow$TD500]     & \ding{52} & 85.9      & 82.8      & 84.3 \\
Ours[15VID$\rightarrow$TD500]+AUG & \ding{52} & 87.9      & \bf{83.1} & \bf{85.4} \\
\hline
\end{tabular}
\end{center}
\end{table}

\subsection{Comparison with Cross Domain and Data Generation Methods}

In this section, we compare our method with some cross-domain and data generation text detection methods. The specific results are shown in Table~\ref{tab:generating}, 15VID-2 represents the results of the second iteration in Table~\ref{tab:iter}. ``A + B'' means the detector trained with A and B together and ``A$\rightarrow$B'' means the detector trained on A and then fine-tuned on B. Without using the data of the target domain, the performance of f-measure increases by 11\% on IC15. \cite{zhan2019ga} uses the domain adaptation network and requires annotations to pre-cut the text area, which is unfair to compare our work with directly. Even though, our performance on IC15 is comparable with theirs. With annotated data in the target domain, the f-measure is increased by 3.2\%. Our method can surpass the domain adaptation network because our method can get more abundant and realistic samples than the generated ones.

\subsection{Comparison with SOTA Methods}

In this section, we compare our method with some state-of-the-art (SOTA) text detection methods. On the IC15 dataset, we will first pre-train the detector on 15VID, and then fine-tune 20 epochs on the IC15 dataset. When evaluating, we will set the long side of the image to 1800. When using the extra data set, we will use 15VID and MLT17 for pre-training. The experimental results are shown in Table~\ref{tab:compare_15}. It is worth noting our detector (Mask R-CNN) has no modification and our tracker (template matching) is very simple while other SOTA models are more sophisticated. We can achieve comparable results with SOTA only by self-training. Moreover, our performance without extra dataset exceeds some methods which use the 800k images \cite{gupta2016synthetic} as an extra dataset. This shows that the samples we mine through the self-training framework are of great value.
On TD500, we follow the settings on IC15 and only set the long side of the test image to 1333. The results are shown in Table~\ref{tab:compare_td}. There is a huge gap in annotation granularity between 15VID and TD500, but our detector can still improve its performance through self-training.


\section{Conclusion and Future Work}
In this paper, we propose a simple yet effective self-training framework for domain adaptive scene text detection. With some sophisticated techniques including TMM, Gen-Loop and balance loss, our method is superior to data generation methods and comparable with the previous domain adaptation method. With real data contained, the state-of-the-art performance can be achieved. For future work, more advanced detectors such as PSENet or SPCNet and more precise tracking techniques such as TrackR-CNN can be used to improve the self-training performance.


{\small
\bibliographystyle{IEEEtran}
\bibliography{ref}
}

\end{document}